%% file: main.tex
\documentclass[english]{lni}

\input{sections/preamble.tex}
\setboolean{anonymous}{false}

\ifthenelse{\boolean{anonymous}}{\author{Anonymous submission}}{
\author{\vspace{-0.5cm} Laurent Colbois\footnote{\emph{Biometrics Security \& Privacy}, Idiap Research Institute~$\vert$~\emph{École des Sciences Criminelles}, University of Lausanne~$\vert$~Switzerland~$\vert$~laurent.colbois@idiap.ch}, Sébastien Marcel\footnote{\emph{Idem}~$\vert$~sebastien.marcel@idiap.ch}}}
\title{On the detection of morphing attacks generated by GANs}

\begin{document}

\maketitle

\renewcommand{\refname}{References}
\setcounter{footnote}{2} 
\thispagestyle{titlepage}
\pagestyle{fancy}
\fancyhead{} 
\fancyhead[RO]{\small Deep-MAD \hspace{25pt}  \hspace{0.05cm}}
\fancyhead[LE]{\hspace{0.05cm}\small  \hspace{25pt}
    \ifthenelse{\boolean{anonymous}}{Anonymous submission}{Laurent Colbois and Sébastien Marcel}
}

\fancyfoot{} 
\renewcommand{\headrulewidth}{0.4pt} 
\vspace{-1cm}
\input{sections/abstract.tex}

\vspace{-0.5cm}
\input{sections/introduction.tex}

\vspace{-0.5cm}
\input{sections/methodology.tex}

\vspace{-0.5cm}
\input{sections/results.tex}
\vspace{-0.5cm}
\input{sections/conclusion.tex}
\vspace{-0.5cm}
\bibliographystyle{lnig}
{\footnotesize
    \bibliography{egbib}
}

\end{document}

%% file: sections/preamble.tex
\IfFileExists{latin1.sty}{\usepackage{latin1}}{\usepackage{isolatin1}}

\usepackage{graphicx}
\usepackage{fancyhdr}
\usepackage{listings} 
\usepackage{changepage} 
\usepackage[figurename=Fig., tablename=Tab., small]{caption}[2008/04/01]

\fancypagestyle{titlepage}{
\fancyhead[RO]{\small A. Br\"omme,  N. Damer,  M. Gomez-Barrero,  K. Raja,  C. Rathgeb, \linebreak A.  Sequeira,  M. Todisco,  and A. Uhl (Eds.): BIOSIG 2022, \linebreak Lecture Notes in Informatics (LNI), Gesellschaft f\"ur Informatik, Bonn 2022} 
\fancyfoot{}}

\renewcommand{\headrulewidth}{0.4pt} 
\usepackage{times}
\usepackage{epsfig}
\usepackage{graphicx}
\usepackage{amsmath}
\usepackage{amssymb}
\usepackage{subcaption}
\usepackage{comment}
\usepackage{multirow}
\usepackage{wrapfig}
\usepackage{ifthen}

\usepackage{todonotes}
\usepackage{diagbox}
\usepackage{microtype}
\usepackage{anyfontsize}
\usepackage{bm}

\usepackage[pagebackref=true,breaklinks=true,colorlinks,bookmarks=false]{hyperref}

\newboolean{anonymous}

\DeclareMathAlphabet{\mathcal}{OMS}{cmsy}{m}{n}

%% file: sections/abstract.tex
\begin{abstract}
Recent works have demonstrated the feasibility of GAN-based morphing attacks that reach similar success rates as more traditional landmark-based methods. This new type of ``deep'' morphs might require the development of new adequate detectors to protect face recognition systems.
We explore simple deep morph detection baselines based on spectral features and LBP histograms features, as well as on CNN models, both in the intra-dataset and cross-dataset case. We observe that simple LBP-based systems are already quite accurate in the intra-dataset setting, but struggle with generalization, a phenomenon that is partially mitigated by fusing together several of those systems at score-level. We conclude that a pretrained ResNet effective for GAN image detection is the most effective overall, reaching close to perfect accuracy. We note however that LBP-based systems maintain a level of interest : additionally to their lower computational requirements and increased interpretability with respect to CNNs, LBP+ResNet fusions sometimes also showcase increased performance versus ResNet-only, hinting that LBP-based systems can focus on meaningful signal that is not necessarily picked up by the CNN detector.
\end{abstract}

%% file: sections/introduction.tex
\section{Introduction}
\vspace{-0.3cm}
\begin{figure}[htb]
	\centering
	\includegraphics[width=\linewidth]{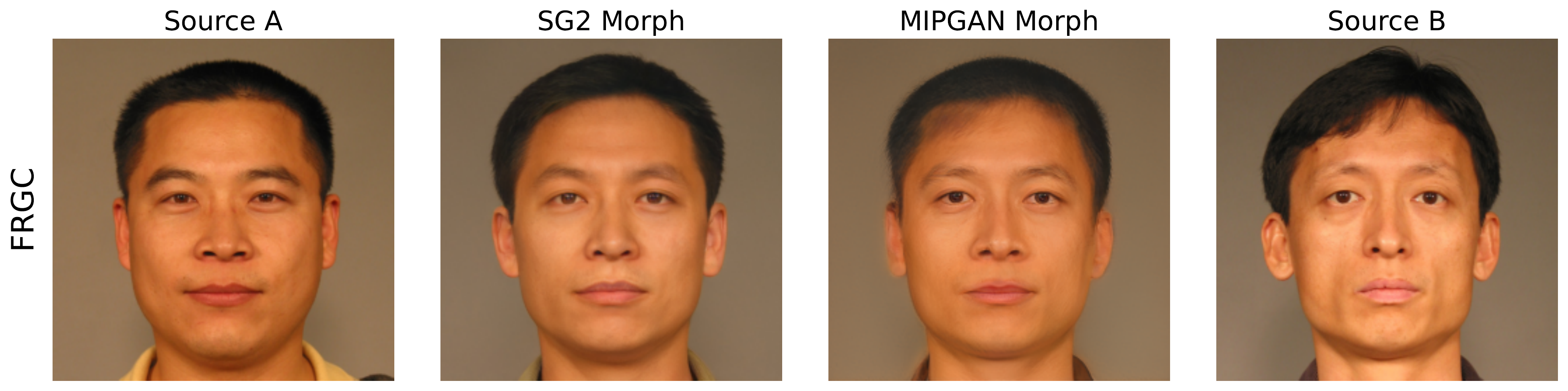}
	\caption{Example of generated deep morphs, using the SG2 method (second image) and the MIPGAN method (third image)}
	\label{fig:morph_examples}
\end{figure}

With the increasing use of automated face recognition (FR) systems, in particular for border control, concern has been raised regarding the vulnerability of such systems to various type of attacks, among which \textbf{morphing attacks}. Morphing attacks enable two individuals to register into the system under the same identity, by providing a fake picture mixing their two faces (the so-called \textbf{morph}) as the reference image at registration time.
The original morphing methodology introduced in \cite{ferraraMagicPassport2014a} (and subsequent variants) is called \textbf{landmark-based morphing} and is based on aligning the landmarks of source images of both merged identities using face warping methods, enabling to create the final morph by simple pixel averaging. This approach has been shown to actually fool common face recognition systems, which has led to the development of various efficient morph detection algorithms \cite{venkateshFaceMorphingAttack2021}, based typically on texture cues, image quality cues, residual noise cues, or using deep neural network (DNN) classifiers.
A new trend in morph generation (\textbf{deep morphing}) has however appeared around 2020, exploiting the quick progress in the field of synthetic face generation using GANs, and in particular the StyleGAN architecture \cite{karrasStyleBasedGeneratorArchitecture2019}. In this context, morphs are obtained by guided exploration of the GAN's latent space, aiming to find a generated image that satisfies high enough identity proximity with respect to both source images. This new type of morphing attack has been shown to also cause vulnerability of common FR systems, albeit to a somewhat smaller extent than landmark-based morphing attacks \cite{zhangMIPGANGeneratingStrong2021a}. Due to the fundamentally different nature of this new generation process, it is unlikely that traditional morph detectors could apply out-of-the-box, and there is to the best of our knowledge still no work focusing specifically on deep morphing attack detection (deep-MAD). The aim of this work is thus to fill this gap by developing and evaluating several deep-MAD systems.  Importantly, literature on generic GAN-image detection has potential to apply. Previous works observe that the convolutional architecture of the generators themselves (and in particular
the upsampling steps) causes them to produce images whose spectral distributions
do not accurately match those observed in real data \cite{durallWatchYourUpConvolution2020}. The specific distribution of a particular network forms what is called a fingerprint of the network, which has potential to be
detected. This is for example exploited by \cite{durallUnmaskingDeepFakesSimple2020} through the use of simple spectral analysis features. To the best of our knowledge, the current state-of-the-art in this domain is the CNN detection network proposed by \cite{wangCNNGeneratedImagesAre2020}, which in particular showcases a strong generalization capability to new GAN architectures.
It is important to note that transferability of those works to the task of deep-MAD is non-trivial and has to be evaluated, due to the way deep morph generation characteristically differs from a simple random sampling of the GAN. Indeed, current algorithms generate deep morphs by exploration of the latent space, usually through optimization and/or interpolation. Moreover, it is common to exploit an extended latent space such as StyleGAN's $\mathcal{W}+$ space introduced in \cite{abdalImage2StyleGANHowEmbed2019}, which is used in every generation method considered in this work. For those two reasons, it is unclear whether there might be a mismatch between the standard output distribution of images produced by the GAN, and the distribution of generated morphs. The morphs could potentially showcase characteristics different enough from straightforwardly sampled images to significantly alter the performance of detectors.  

We will first establish baselines using non DNN-based systems, under the form of simple linear classifiers exploiting two types of handcrafted features, specifically spectral-based by working in the Fourier domain, and texture-based using histograms of Local Binary Patterns (LBPs). Indeed, the difficulty of the detection task is unclear a priori and we find appropriate to approach the problem with simple systems first. Moreover, the available data is quite scarce~: most morphing datasets appearing in the literature are in the order of magnitude of 1K images, and they are typically not publically available, which might limit the capability of training DNNs for deep-MAD. Finally, the higher interpretability of handcrafted features with respect to DNN systems maintain their relevance in certain applications, such as the analysis of a single suspected morph in a forensic context, where explanations beyond the pure system prediction are typically required. In a second stage, we will further experiment with 4 CNN architectures trained specifically for deep morph detection, and finally with a CNN pretrained for GAN image detection that we reuse out-of-the-box with no fine-tuning. We will finally evaluate whether performance improvements can be obtained by simple score-level fusion of those different approaches.

To summarize, our main contribution is the following: we establish an overview of deep-MAD capability, in particular evaluating how much the problem is similar to the one of GAN-image detection. Note that we choose to exclude landmark-based morphs from our experimentation; this remains compatible with a real-world scenario as joint detection of both landmark-based and deep morphs is not necessarily needed (two independent detection systems could be used).

%% file: sections/methodology.tex
\section{Methodology}
\vspace{-0.3cm}
\paragraph{Datasets} Unfortunately the near-totality of face morphing datasets appearing in the literature are not publically available. For this reason, we generate our own for the purpose of our experiments. We consider two different generation approaches. The \textbf{SG2} method follows the approach presented in \cite{Sarkar_ICASSP_2022} (interpolation between the projection of the two source images in the latent space of a pretrained StyleGAN2 network), with the main difference that we perform this operation in the $\mathcal{W+}$ space of the GAN, introduced in \cite{abdalImage2StyleGANHowEmbed2019}. The \textbf{MIPGAN} method is the one introduced in \cite{zhangMIPGANGeneratingStrong2021a}, a derivation of the SG2 which includes a biometric loss during the latent space exploration, thus providing higher guarantees that the resulting synthetic images act effectively as morphs.

We sample the source identities from two datasets : the Face Research Lab London Set \cite{debruineFaceResearchLab2017a} (FRLL), and the Face Recognition Grand Challenge 2.0 set \cite{phillipsOverviewFaceRecognition2005} (FRGC). For the FRLL dataset, we use the same pairs of sources as in the AMSL dataset, a landmark-based morphing dataset also based on FRLL and using the morphing method introduced in \cite{neubertExtendedStirTraceBenchmarking2018a}. For the FRGC dataset, we use the same pairs of sources as in \cite{zhangMIPGANGeneratingStrong2021a}. We generate morphs from the FRLL dataset using the SG2 method (FRLL-SG2 dataset), and morphs from the FRGC dataset using both the SG2 and MIPGAN methods (FRGC-SG2 and FRGC-MIPGAN datasets). Figure \ref{fig:morph_examples} showcases examples of the generated morphs. Protocols for reproducing those morphs datasets, as well as code for the \textbf{SG2} method, are released jointly with this article.\footnote{\url{https://gitlab.idiap.ch/bob/bob.paper.biosig2022_deep_mad}} For the code of \textbf{MIPGAN}, we suggest contacting the authors of \cite{zhangMIPGANGeneratingStrong2021a}.

\vspace{-0.5cm}\paragraph{Detection algorithms}
We choose to explore single image detection rather than differential detection (where the morph is compared to a trusted bona fide image). In a quick qualitative analysis of generated deep morphs, we notice in particular that those tend to appear smoother than real images; skin irregularities such as moles vanish, thin hair details disappear, and the skin looks overall a bit more ``plastic''. We hypothesize that deep morphs tend to contain reduced high-frequency content with respect to real images, and that texture-related features might be successful for deep-MAD. We first experiment with \textbf{handcrafted features} that are subsequently used to train a simple linear classifier, specifically a Linear Discriminant Analysis (LDA). \textbf{Fourier features} are first considered : we analyze the frequency content of the images by computing their Fourier transform, then integrating it over radial frequency bands such as proposed in \cite{durallUnmaskingDeepFakesSimple2020} for Face Swap detection. We also experiment with \textbf{Local Binary Patterns histogram features} (LBP) used in the past for face anti-spoofing \cite{chingovskaEffectivenessLocalBinary2012}, that work as local texture extractors. We use the implementation proposed in the Bob toolkit\footnote{\url{https://www.idiap.ch/software/bob/}} and experiment with several configuration variants, differing by the size of the neighborhood ($(n, r)=(4, 1)$, $(8, 1)$ or $(8, 2)$ with $n$ the number of neighbors and $r$ the radius of the neighborhood), the shape of the neighborhood (square or circular), and the optional regrouping of similar codes in the same histogram bins (\emph{RIU2} regrouping where local textures equivalent up to a rotation are grouped in the same bin), leading to a total of 12 different LBP configurations.  In a second stage, we also evaluate the efficiency of \textbf{deep features} by training various CNN architectures for detection. We explore 5 different architectures: two of them are the XCeption and EfficientB4 models that have been successful for deepfake detection in \cite{korshunovImprovingGeneralizationDeepfake2021}. We also experiment with 2 smaller scale architectures, specifically MobileNet and MobileNetV2\footnote{\url{https://www.tensorflow.org/api_docs/python/tf/keras/applications/mobilenet/MobileNet}}, in order to mitigate the risk of overfitting given the small size of the datasets. Training process follows the one presented in \cite{korshunovImprovingGeneralizationDeepfake2021}. For the last architecture, we experiment with a pretrained ResNet50-ProGAN, which is the GAN-image detection model proposed in \cite{wangCNNGeneratedImagesAre2020}. It has potential to show good performance in our situation given its strong generalization capability and the fact that our morphs are themselves produced by GANs. We specifically use the Blur+JPEG(0.1) version that is available for download on GitHub\footnote{\url{https://github.com/peterwang512/CNNDetection}}. Source code to reproduce the experiments is available online.\footnote{\url{https://gitlab.idiap.ch/bob/bob.paper.biosig2022_deep_mad}}

%% file: sections/results.tex
\section{Results}
\vspace{-0.3cm}

We aim to evaluate the effectiveness of the deep-MAD systems both in the intra-dataset and cross-dataset cases. We approach the task as a binary classification problem, where
images belonging to the `morph' class are taken from respectively FRLL-SG2, FRGC-SG2 and FRGC-MIPGAN, and images belonging to the `bona fide' class are taken from the corresponding source datasets.
In the intra-dataset case, we create separate train and development sets by splitting identities of the source datasets in
two groups, 3/4 of them for train and 1/4 for development. This makes the split straightforward for bona fide examples; for morph examples, we include in each set only morphs for which both source identities are also part of the set. Due to the small scale of our datasets, further reduced by the train/dev split, evaluation of CNN approaches is unpractical in the intra-dataset case and is thus reserved for the cross-dataset case.
In the cross-dataset case (evaluation of the generalization capability), we systematically train on each (full) dataset and evaluate on the two others. We note that FRLL-SG2 and FRGC-SG2 use the exact same GAN (architecture and weights), and the same morph generation algorithm, but use different source datasets.
FRGC-SG2 and FRGC-MIPGAN use the same source dataset and the same GAN architecture, but a different set of weights (for FRGC-MIPGAN the GAN is fine-tuned on the FRGC dataset) and a different morph generation algorithm. We also report results of the ResNet50 network trained on ProGAN image for GAN-image detection, that we reuse out-of-the-box.

We choose to focus on summary metrics rather than on specific operating point performances, mainly the area under the AUC ($\in [0, 1]$, higher is better) which is popular in deepfake detection literature (e.g. \cite{korshunovImprovingGeneralizationDeepfake2021}), and the EER $\%$ ($\in [0, 100]$, lower is better). We report only the best performing configuration for each type of system. Intra-dataset and cross-dataset results are presented in tables \ref{tab:mad}, resp. \ref{tab:crossdbmad}.

\begin{table}[htb]
	\footnotesize
	\centering
	\begin{tabular}{|lll|rr|}
		\hline
		Dataset & Extractor & Specs            & {AUC}          & {EER (\%)}     \\
		\hline\hline
		FRLL-   & Fourier   &                  & 0.92           & 17.31          \\
		SG2     & LBP       & (8,2) $\bigcirc$ & \bfseries 0.99 & \bfseries 5.77 \\
		\hline
		FRGC-   & Fourier   &                  & 0.92           & 16.21          \\
		SG2     & LBP       & (8,2) $\bigcirc$ & \bfseries 1.00 & \bfseries 2.86 \\
		\hline
		FRGC-   & Fourier   &                  & 1.00           & 1.43           \\
		MIPGAN  & LBP       & (8,1) $\square$  & \bfseries 1.00 & \bfseries 0.00 \\
		\hline
	\end{tabular}
	\caption{MAD performance on the development sets of our three morphing attack datasets. For LBP features, we only report the LBP configuration with the maximum AUC. $\bigcirc$ indicates a circular LBP shape while $\square$ indicates a square LBP shape.}
	\label{tab:mad}
	\vspace{-0.3cm}
\end{table}

We observe than despite the simplicity of the classifiers, both types of features enable significantly much better than random detection capability, but that LBP-LDA systems always outperform Fourier-based ones. We observe in particular that MIPGAN morphs seem much easier to detect, with the Fourier-LDA system performing significantly better than on SG2-morphs, and the (8,1) $\square$ LBP-LDA system even being able to perform perfect separation. This is to be put in contrast with the fact that MIPGAN morphs usually cause higher vulnerability of common FR systems than SG2 morphs. We do note that visually, MIPGAN morphs seem more regularly to contain visible artifacts, especially at the level of the hair, which sometimes tends to look a bit blurry. This blurriness might be transcribed as a more characteristic spectral distribution, hence facilitating detection.

\begin{table}[htb]
	\fontsize{6.5}{8}
	\selectfont
	\centering
	\begin{tabular}{|l|lll|rr|}
		\hline
		{Train on} & {Test on}   & {Extractor} & {Specs}               & {AUC}          & {EER (\%)}      \\

		\hline\hline
		FRLL-      & FRGC-       & Fourier     &                       & 0.88           & 20.46           \\
		SG2        & SG2         & LBP         & RIU2 (8,1) $\square$  & 0.93           & 15.16           \\
		           &             & CNN         & Xception              & \bfseries 0.99 & \bfseries 5.72  \\
		\cline{2-6}
		           & FRGC-       & Fourier     &                       & 0.45           & 53.25           \\
		           & MIPGAN      & LBP         & RIU2 (8,2) $\square$  & 0.96           & 11.39           \\
		           &             & CNN         & Xception              & \bfseries 0.99 & \bfseries 5.72  \\
		\hline
		FRGC-      & FRLL-       & Fourier     &                       & 0.92           & 16.04           \\
		SG2        & SG2         & LBP         & RIU2 (8,1) $\square$  & 0.91           & 19.97           \\
		           &             & CNN         & EfficientNet          & \bfseries 1.00 & \bfseries 1.20  \\
		\cline{2-6}
		           & FRGC-       & Fourier     &                       & 0.85           & 23.24           \\
		           & MIPGAN      & LBP         & (8,2) $\bigcirc$      & 0.99           & 3.91            \\
		           &             & CNN         & EfficientNet          & \bfseries 1.00 & \bfseries 0.00  \\
		\hline
		FRGC-      & FRLL-       & Fourier     &                       & 0.62           & 41.8            \\
		MIPGAN     & SG2         & LBP         & RIU2 (8,2) $\bigcirc$ & \bfseries 0.92 & \bfseries 16.98 \\
		           &             & CNN         & MobileNet             & 0.89           & 19.2            \\
		\cline{2-6}
		           & FRGC-       & Fourier     &                       & 0.79           & 28.51           \\
		           & SG2         & LBP         & (8,2) $\bigcirc$      & 0.97           & 8.76            \\
		           &             & CNN         & MobileNet             & \bfseries 0.99 & \bfseries 5.41  \\
		\hline
		ProGAN     & FRLL-SG2    & CNN         & ResNet50              & \bfseries 1.00 & \bfseries 0.00  \\
		           & FRGC-SG2    & CNN         & ResNet50              & \bfseries 1.00 & \bfseries 0.56  \\
		           & FRGC-MIPGAN & CNN         & ResNet50              & \bfseries 1.00 & \bfseries 0.00  \\
		\hline
	\end{tabular}

	\caption{Cross-dataset MAD performance on our three morphing attack datasets. For LBP features and CNNs, we pick the best performing configuration on the development set and report performance on the evaluation set. $\square$ indicates a square LBP shape, while $\bigcirc$ indicates a circular shape.}
	\label{tab:crossdbmad}
\end{table}
Moving to the cross-dataset setting, we observe a significant decrease in detection performance of the LDA classifiers. LBP-LDA classifiers still showcase better performance than Fourier-LDA ones in almost every case (often significantly). However, we note the cross-dataset generalization error they showcase is often more important. For example, alternating between FRLL-SG2 and FRGC-SG2 causes very limited performance decrease with Fourier-LDA, whereas the EER is multiplied by around 3x-6x with LBP-LDA. A similar phenomenon is observed when alternating between FRGC-SG2 and FRGC-MIPGAN, except for the Fourier-LDA generalization error being multiplied by $\sim 20$ going from FRGC-MIPGAN to FRGC-SG2; in this case this is probably an additional showcase of the ``surprisingly'' high effectiveness of Fourier feature on FRGC-MIPGAN observed in the intra-dataset experiment. Finally, alternating between FRLL-SG2 and FRGC-MIPGAN, Fourier-LDA classifiers perform very poorly, while LBP-LDA showcases similar generalization error than in the other cases. Overall, it seems the LBP-LDA features are more \emph{consistently} robust than Fourier-LDA ones, which are relatively robust when keeping the same morphing algorithm, but not really otherwise (and perform worse than LBP features overall).

We note however that no LBP configuration emerges as the systematic best choice. Overall, rotation-invariant regrouping and 8-neighbors shapes generally correlate with good performance, but we do observe that no bins regrouping is the best setup for FRGC-SG2 / FRGC-MIPGAN generalization. In this latter case, we hypothesize that this might be an effect of the shared bona fide set between both datasets : not using bins regrouping decreases regularization and could enable the classifier to learn more patterns specific to this bona fide set, thus helping with generalization.

The CNN classifiers showcase drastically better cross-dataset performance, significantly outperforming LBP-LDA in all cases except when training on FRGC-MIPGAN and testing on FRLL-SG2. In this specific case, it does seem more about the CNN underperforming rather than LBP-LDA performing very well. Interestingly, this is not true in the reverse case (training on FRLL-SG2 and testing on FRGC-MIPGAN), suggesting that features learned by the CNN on SG2-based datasets might generalize better to the MIPGAN-based dataset, rather than the reverse.

Last but not least, the ResNet50-ProGAN system beats every other classifier by a large margin, despite having been used out-of-the-box with no fine-tuning.
This shows the generalization power of this GAN-image detector is preserved despite replacing a simple image sampling procedure by a much more complex $\mathcal{W}+$ latent space exploration process Again, the preservation of this generalization capacity was not an a priori guarantee.

It thus looks like simply reusing a GAN-image detector is currently the most effective way to approach deep-MAD. However, we remain interested in exploring ways to push the performance of classifiers based on handcrafted features, which have other advantages such as lower computational cost, and higher interpretability. Having already access to a large quantity of classifiers potentially focusing on complementary aspects of the data, we thus further experiment with score-level fusion of those classifiers. This is achieved considering pairings of independent systems and training a linear regression classifiers on one of the datasets (the calibration set), using the two set of scores as features. Performance on the remaining datasets is then evaluated. We consider two scenarios : fusion between two LDA systems (\emph{can we improve performance while still only considering simple systems ?}) and fusion between an LBP-LDA system and the ResNet system (\emph{can an LBP-LDA system still help to push the performance of the ResNet ?}). Results are reported in table \ref{tab:fusion}.

We observe that this fusion process almost systematically leads to performance improvements, sometimes significant for the LDA systems. More interestingly, it also improves performance in some cases (e.g. testing on FRGC-SG2) over using the standalone ResNet. This improvement is small (EER decrease of around 0.4\%) but non-negligible, corresponding to a decrease from 11 to 3 erroneous classifications over 1940 queries. This suggests the LDA systems still provide complementarity to the ResNet, despite the significantly better single-system performance system of the latter. On more challenging datasets especially, simple handcrafted features could thus still be of use to improve performance beyond the single-CNN one.

\begin{table}
	\fontsize{6.5}{8}
	\selectfont
	\centering
	\begin{tabular}{|l|llc|cc|}
		\hline
		Calib. on & Test on & Specs      & Fused EER (\%) & Sys. 0 EER (\%) & Sys. 1 EER (\%) \\
		\hline
		FRLL-     & FRGC-   & All LDA    & 10.00          & 15.16           & 20.46           \\
		SG2       & SG2     & LBP-LDA+ResNet & \bfseries 0.15 & 17.94           & 0.56            \\
		\cline{2-6}
		          & FRGC-   & All LDA    & 10.98          & 11.39           & 13.09           \\
		          & MIPGAN  & LBP-LDA+ResNet & \bfseries 0.00 & 39.23           & 0.00            \\
		\hline
		FRGC-     & FRLL-   & All LDA    & 12.03          & 19.97           & 16.04           \\
		SG2       & SG2     & LBP-LDA+ResNet & \bfseries 0.00 & 20.99           & 0.00            \\
		\cline{2-6}
		          & FRGC-   & All LDA    & 3.65           & 3.91            & 7.90            \\
		          & MIPGAN  & LBP-LDA+ResNet & \bfseries 0.00 & 26.03           & 0.00            \\
		\hline
		FRGC-     & FRLL-   & All LDA    & 15.02          & 16.98           & 20.99           \\
		MIPGAN    & SG2     & LBP-LDA+ResNet & 0.76 & 19.03           &  \bfseries 0.00            \\
		\cline{2-6}
		          & FRGC-   & All LDA    & 8.09           & 8.76            & 21.44           \\
		          & SG2     & LBP-LDA+ResNet & \bfseries 0.15 & 10.98           & 0.56            \\
		\hline
	\end{tabular}

	\caption{Cross dataset MAD performance using linear logistic regression score-level fusion of 2 systems. Among all considered combinations, we pick the best performing one on the development set and report performance on the evaluation set. For LDA systems, the same data is used for training \& calibration, while the ResNet system is trained on ProGAN. In the LDA+ResNet case, system 1 always refers to the ResNet system.}
	\label{tab:fusion}
\end{table}

%% file: sections/conclusion.tex
\section{Conclusion}
We have showcased the performance of several systems for deep-MAD. LBP-LDA approaches have reasonable merit in the intra-dataset case, but suffer from an important generalization error in the cross-dataset setting. We observe that CNN-based approaches typically perform better. In particular, the ResNet50-ProGAN model, originally trained for robust GAN-image detection, still performs very well here. This seems to confirm the problematic of deep-MAD is strongly equivalent to the one of GAN-image detection, a hypothesis which was not necessarily trivial given the way deep morphs are obtained through a latent space exploration process, rather than through simple GAN sampling. Finally, we observe that additionally to their advantages of low computational requirements and better interpretability, simple LBP-LDA systems can still provide complementary information to the ResNet system, leading to improved performance when fusing the two.

We identify two main directions to extend this work. First, an extended study considering Print-Scan post-processing of the images is necessary to better simulate a real-world scenario. Secondly, an extension should focus on joint detection of both landmark-based morphs and deep morphs : a real-world scenario could include both types of attacks, thus ideally requiring a system performing well in both cases. Noting that LBP have in the past been successful for landmark-based morphs detection \cite{venkateshFaceMorphingAttack2021}, the systems presented in this work have potential to be exploitable in this joint setup.

\ifthenelse{\boolean{anonymous}}{}{
    \vspace{-0.5cm}
\paragraph{Acknowledgements}
This work was supported by the Swiss Center for Biometrics Research \& Testing and the Idiap Research Institute.
}